# Prometheus Chatbot: Knowledge Graph Collaborative Large Language Model for Computer Components Recommendation

Songhao Chen[1,#,*], Yunsheng Wang[2,#] and Kevin Jin[3]

[1]Lenovo Desktop Computing Development Lab, Shenzhen, China

[2]McKelvey School of Engineering, Washington University in St. Louis, St. Louis, MO, USA

[3]Lenovo Desktop Computing Development Lab, Beijing, China

[#]These Authors contributed equally to this work

[*]Corresponding author: chensh8@lenovo.com

**Abstract**. Knowledge graphs (KGs) are essential in applications such as network alignment, question-answering, and recommender systems (RSs) since they offer structured relational data that facilitate the inference of indirect relationships. However, the development of KG-based RSs capable of processing user inputs in natural language faces significant challenges. Firstly, natural language processing units must effectively handle the ambiguity and variability in human language to interpret user intents accurately. Secondly, the system must precisely identify and link entities, like product names, to their corresponding nodes in KGs. To overcome these challenges, supported by Lenovo, we developed a novel chatbot "Prometheus" integrating KG with a large language model (LLM), specifically used for recommending computer components. This chatbot can accurately decode user requests and deliver personalized recommendations derived from KGs, ensuring precise comprehension and response to their computer setup needs.

## 1. Introduction

Knowledge Graph (KG) is a graph structure that contains a collection of facts, where nodes represent real-world entities, events and objects, and edges denote the relationship between two nodes (Fensel et al., 2020). Since its debut in 2012, various KGs have been generated, including Freebase, Yago, Wikidata and so on (Färber et al., 2018). The applications of KGs are numerous, ranging from network alignment, and question-answering to recommender systems (RSs). Among these applications, KG-based RSs aim to process user inputs and provide related recommendations.

The core of any RS is its personalized recommendation algorithm. Traditional algorithms have achieved notable success and are generally divided into content-based filtering (Van Meteren et al., 2000), collaborative filtering (Herlocker et al., 2000), and hybrid filtering techniques (Basilico et al., 2004). Despite their successes, these traditional methods still face unresolved issues, such as sparse relationships between users and items and the cold-start problem (Lika et al., 2014) encountered when recommending to new users. Additionally, scaling these algorithms to meet the demands of real-world recommendation scenarios presents a significant challenge. To address these limitations, researchers and engineers have incorporated auxiliary information into the RSs, including attribute characteristics of users and items (Wang et al., 2018, Wang et al., 2018, Wang et al., 2018), user social network information (Jamali & Ester, 2010), and multimedia information (e.g., texts (Wang et al., 2015), images (Zhang et al., 2016)). Notably, KGs offer a wealth of background information on items, revealing hidden relationships between them, which significantly enhances recommendation quality (Ji et al., 2021).

Recently, RSs based on KGs have gained attention from both academia and industry, leading to the development in numerous domains (Chicaiza et al., 2021). In the education domain, RSs are now a critical area of research (Manouselis et al., 2011), with a particular focus on aiding learners by identifying educational resources (Chicaiza et al., 2017) and academic content suitable for integration into their studies (Jia et al., 2018; Deng et al., 2019; Li et al., 2019). Similarly, in the tourism industry, these RSs provide tailored information, enhancing user experience (Guo et al., 2019; Nilashi et al., 2015; Santhoshi et al., 2020; Wang et al., 2019; Wu et al., 2020; Suzuki et al., 2019). KGs are also increasingly utilized in health (Gim et al., 2018; Wang et al., 2020), entertainment (Catherine et al., 2017), and business (Wang et al., 2019; Hu et al., 2019).

KG-based RSs generally consist of KGs, a recommendation module, and a connection module. Among them, KGs store rich semantic information about entities, and the recommendation module calculates the user-item relation. The connection module maps the semantic information into a low-dimensional vector and then combines it with the recommendation module to calculate the interaction information between users and items. Unlike the traditional connection model, we use a Large Language Model (LLM) to process user inputs.

In this paper, supported by Lenovo, we explore the creation of a unique KG-based RS named "Prometheus" chatbot for recommending computer components. The primary goal of this project is to simplify the interaction with documents and extract valuable information using natural language. Prometheus chatbot is built using LangChain and GPT-4/ChatGPT to deliver a smooth and natural conversational experience to the user, with support from both Azure OpenAI Services and OpenAI. The key contributions of our work are as follows:

We developed KGs of computer components using Lenovo's proprietary database, which includes product names and their related rules.

We combined KGs with an LLM to create the Prometheus chatbot that establishes a new standard in the industry, especially those dealing with complex computer components.

## 2. Proposed Method

### 2.1 Model Overview

The Prometheus chatbot integrates Knowledge Graph Construction, Natural Language Processing, and Chatbot design to create a completed recommender system for computer components. This system is built upon three main pillars: the Neo4j database for constructing the knowledge graphs, the Azure OpenAI service for processing natural language, and the Streamlit for building the Chatbot interface.

### 2.2 Knowledge Graph (KG) Construction

KG construction is an iterative engineering process that involves various methods and tools. These approaches can be categorized into two main clusters: top-down and bottom-up.

The top-down approach is rooted in the modeling processes used in database construction. It begins with identifying a subject domain and a list of research needs, followed by designing a conceptual model to collect the entities of interest, their inter-relationships, and the categories. Tools like CmapTools are valuable for conceptual modeling. This is followed by creating logical and physical models that add logical representation and assertions to the collected entities and relationships. During the technical development and implementation phase, important considerations include the coding language (e.g., RDF and OWL), serialization formats (e.g., RDF/XML, Turtle, and JSON-LD), and KG development platforms such as Protégé and DOGMA. The final step is deploying the KG as a service to allow community reuse and feedback, transforming domain knowledge into a machine-readable representation.

In contrast, the bottom-up approach to KG construction relies on crowd-sourced data, such as social media and legacy literature. The expansion of social media and open access to published literature has significantly increased data sources, leading to a rise in publications using this approach in recent years. Although the bottom-up approach can quickly build large KGs by processing extensive datasets, a challenge remains in the precise logical representation and assertions for the entities and relationships.

In this study, our approach is a combination of top-down and bottom-up methods. We begin with a top-down strategy by identifying specific types of rules (Text-rule, Derive, and Select) from an SQLite database (T3.db) and designing a conceptual model that includes entities of interest and their inter-relationships. This involves logical and physical modeling, where extracted rules are standardized and processed to create nodes and relationships in Neo4j, ensuring a structured and precise representation of data. Simultaneously, we employ a bottom-up approach by extracting raw data through SQL queries and preprocessing it to remove unnecessary elements and standardize formats. This dual-method approach leverages the structured design of the top-down method while incorporating the data-driven extraction and transformation processes characteristic of the bottom-up approach.

The construction of knowledge graphs has been identified as a promising area for enhancing recommender systems. In this study, we use the graph database Neo4j to construct KGs (Guia et al., 2017). A notable benefit of Neo4j is its streamlined approach to identifying connections between data points. While programming languages like Python with py2neo process connections at the time of the query, Neo4j stores the connections directly in the database, significantly enhancing performance. This study utilizes Neo4j's free version, allowing unrestricted replication of our results.

**2.3 Prometheus Chatbot Design**

The Streamlit application serves as a user-friendly front end for interacting with the KGs of computer components (Khorasani et al., 2022). This interface leverages the Azure OpenAI service to understand and process natural language queries, making the system highly intuitive and accessible even for users with limited technical knowledge.

**Features of the Prometheus Chatbot:**

**2.3.1 Natural Language Processing with Azure OpenAI service:**

Large language models (LLMs) demonstrate impressive capabilities in natural language processing, which are constructed using the transformer architecture. These models combine large-scale architectures with huge amounts of textual training data. This scaling up has allowed LLMs to understand and generate text at a level comparable to that of humans. Among them, ChatGPT emerged as the hottest topic on the Internet at the end of 2022 and established itself as a "cultural sensation". In this paper, we use Microsoft Azure OpenAI service to access the ChatGPT API for processing user-natural languages into Cypher queries. For example, users can input questions or queries such as "Please recommend me the power supply about GFX 3050." or "What components are recommended if I buy an intel i7 CPU?" The LLM then interprets these queries and generates appropriate responses by referencing the KG we constructed. This integration enables the system to handle various question formats and complexities, providing users with precise recommendations.

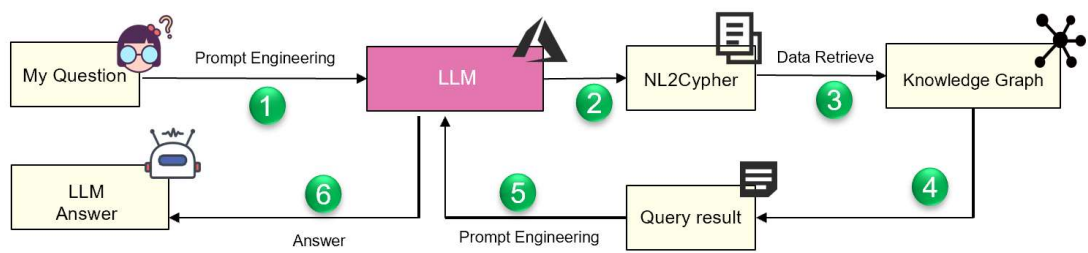

Figure 2. LLM Structure Data Query Architecture. The process starts by injecting a user's question into the system as input. The LLM then interprets this input and generates a Cypher query, which is executed against a specified database or KG. The results of this query are retrieved and fed back into the LLM. This data is processed and integrated into the LLM's prompt, allowing it to generate a coherent and contextually appropriate answer. The final step is the output, where the LLM provides the user's answer in natural language.

**2.3.2 Search Functionality:**

The search process begins with the user inputting a query, such as "Tell me the GFX3050 T3 rule about M70t Gen5." The LLM then extracts relevant keywords from the query, identified here as `['3050', 'M70t Gen5']`. These keywords are injected into the KG. The query seeks nodes whose `name` attribute contains '3050' and whose `project name` attribute contains 'M70t Gen5', returning the `T3 rules` attribute of the matching nodes. Finally, the system retrieves this information from the KG and provides the answer to the user.

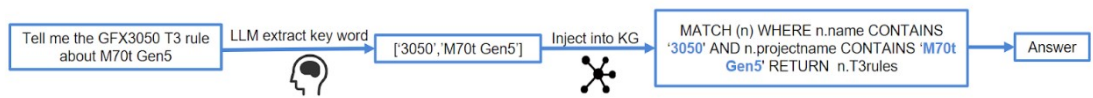

Figure 3. Illustration of the Prometheus chatbot search functionality

**2.4 Workflow of the Prometheus Chatbot:**

1. The user inputs a query in natural language via the Streamlit interface.
2. The LLM processes the input, interprets the user's intent, and formulates a query for the KG.
3. The system queries the Neo4j database to retrieve relevant nodes based on the processed query.
4. The LLM generates a user-friendly response to present the information.
5. The Prometheus Chatbot interface displays the results in tabular and graphical formats, allowing users to explore the data interactively.

Generally, we develop a novel tool that improves the precision of recommendations for computer components. This system simplifies the process of finding compatible components and uses an LLM to offer recommendations that are context-aware and personalized.

**3 Related Work**

**3.1 Data Preprocessing**

This step involves querying an SQLite database named T3.db, a proprietary database owned by Lenovo, to retrieve rules for computer components. We focus on three specific types of rules: Text-rule, Derive, and Select. SQL queries are designed to fetch these rules with their associated metadata, ensuring selection based on relevancy, rule type, and content. This stage is crucial for gathering the raw data to populate our KG.

Figure 4. Excerpt from the raw data in T3.db. The "owner" column is restricted for privacy reasons.

The extracted rules are then processed to remove unnecessary parts and standardize their format. The processing steps include: removing unexpected words and characters from the original rules (such as '||' and '&&'), applying a uniform format for all rules, and dividing complex rules into simpler sections (for complex rules that combine several simpler criteria, such as "( ✔ Intel i5-12400 2.5GHz/6C/12T/18M 65W DDR5 4800 || ✔ Intel i5-12400F 2.5GHz/6C/12T/18M 65W DDR5 4800 || ✔ Intel i5-12500 3.0GHz/6C/12T/18M 65W DDR5 4800)".

### 3.2 Knowledge Graph Construction

The cleaned data is then used to construct a KG using Neo4j. Nodes and relationships are created to represent components and their associated rules respectively. Each node consists of its name, original rule, rule index, type, project name, date, and owner. Relationships between nodes are defined based on the rules, indicating compatibility (should/should not). The Neo4j database is populated with nodes and relationships. Take the "Derive" Type, for example, 32,776 nodes and 24,971 relationships are created.

| Parent Node | Child Node | Relationship |
|---|---|---|
| PCI Card Holder Kit for RTX3050 8G | RTX3050 8GB G6 128b H+3DP HP | should |
| SATA 2TB 7200 RPM/6Gb | Optional 3.5HDD screw and grommet kit | should not |

Table 1. Sample Parent-Child Nodes and Relationships

### 3.3 Prometheus Chatbot Building

The chatbot is essential in connecting users with the RS and understandably presenting complex data. In this section, we use Streamlit to construct the Prometheus chatbot for tabular data representation (Khorasani et al., 2022). This design ensures that users can quickly comprehend and compare different components, making it easier to determine what additional components are necessary for their specific computer setup needs.

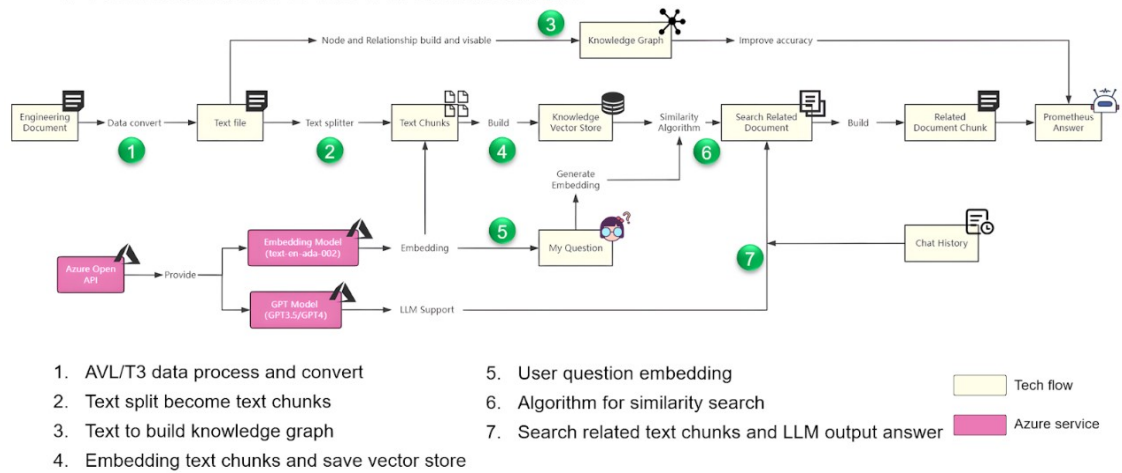

Figure 5. Illustration for the Prometheus V1.0 Architecture. Prometheus chatbot processes engineering documents by converting the data into text files, splitting the text into chunks, and building a KG to visualize relationships and improve accuracy. These text chunks are embedded using an Azure-provided embedding model and stored in a vector store. When a user asks a question, it is embedded and compared against the vector store to find related documents. The system uses an LLM to generate accurate answers based on these documents, providing coherent responses to user queries.

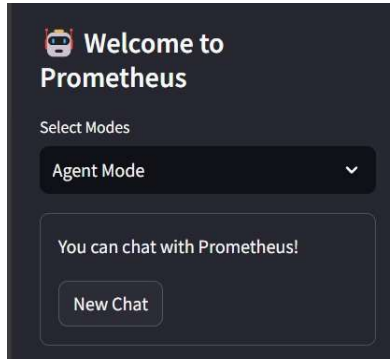

Figure 6. Cover of the Prometheus Chatbot interface. Users can toggle between different agent modes to select the appropriate KG database according to their needs.

## 4 Results

### 4.1 Knowledge Graphs:

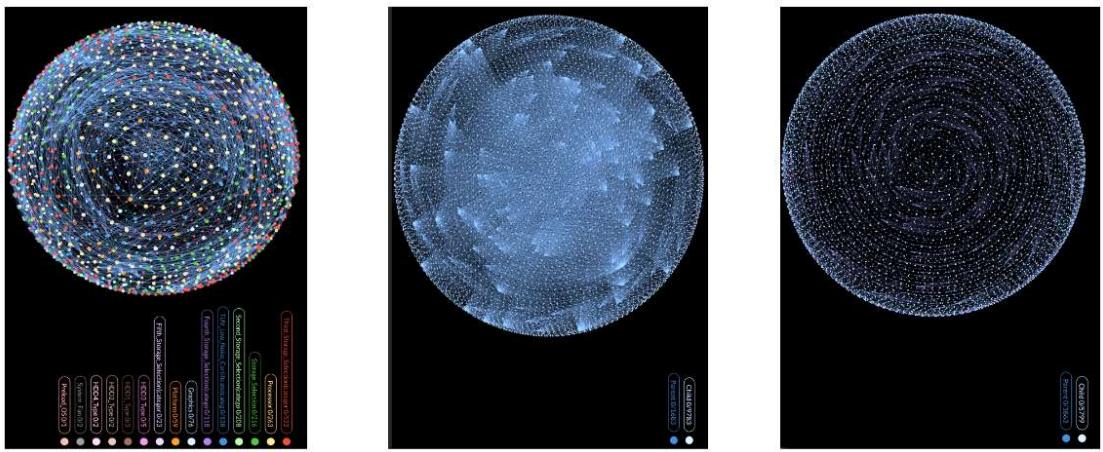

Figure 7. Knowledge Graph Representations for "Text-rule", "Derive" and "Select" Type Datasets. These are three types of constraints in Lenovo desktop computer components: "Text-rule" represents special compatibility rules among main components, "Derive" represents compatibility rules between main and secondary components, and "Select" represents ordinary compatibility rules for main components. In the dataset of the "Text-rule" type, the data facilitates the categorization of nodes by distinct colors. Conversely, for datasets labeled as "Derive" and "Select" types, we use parent and child nodes to illustrate the connections.

### 4.2 Zoom-In Nodes and Relationships:

In KGs, each computer component is represented as a node. Relationships between nodes are presented as edges, which denote should/should not.

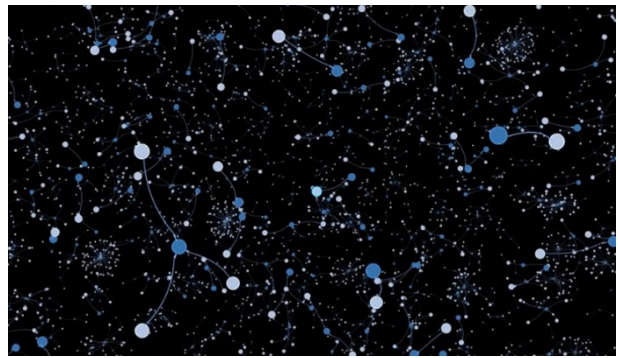

Figure 8. Detailed View of the "Derive" Type Knowledge Graph Section generated by GraphXR

For instance, a Memory Unit '64GB(32+32) DDR4 3200 UDIMM' node has a relationship with its parent node '32GB DDR4 3200 UDIMM' indicating that it is an extension of the base configuration, allowing for increased memory capacity while maintaining compatibility in terms of memory type, speed, and form factor. This structure allows the system to perform checks and suggest optimized component pairings tailored to specific use cases.

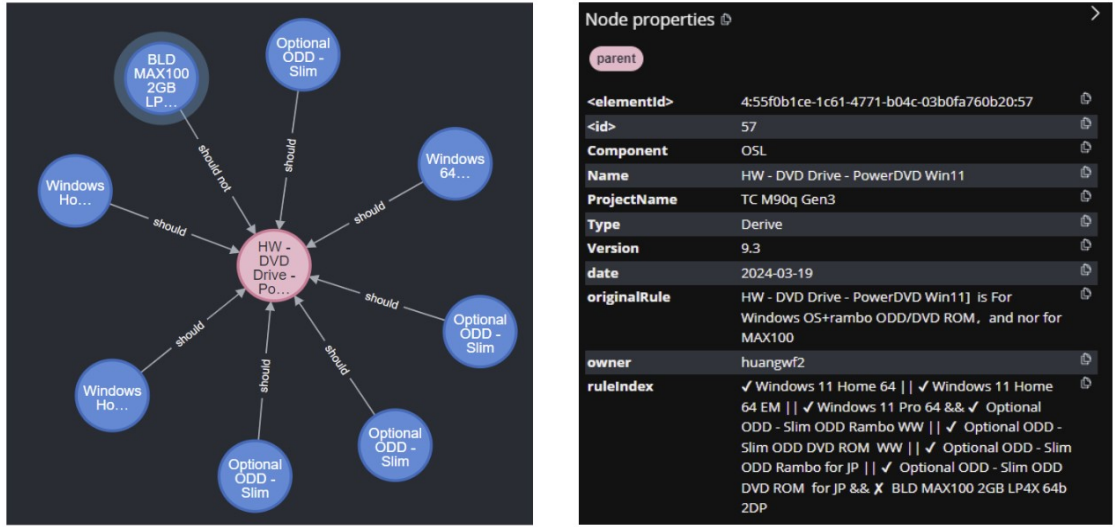

Figure 9. Example of Parent and Child Nodes in the "Derive" Type dataset. The pink node in the center represents the parent node, while the surrounding blue nodes are all child nodes. The right image is the attribute graph of the central parent node.

### 4.3 Prometheus Chatbot Operation:

Search results and other relevant data are presented in a clear and organized table format, allowing users to quickly understand and compare different components.

Figure 10. An example of the Prometheus Chatbot Operation. This figure shows the results from a KG query related to the rules for using the RTX 3050 graphics card in the ThinkCentre M70t Gen 5. The table provides the constraints and requirements for deploying the RTX 3050 graphics card in the specified system, ensuring compatibility and optimal performance.

### 5 Discussion

Our work shows that integrating knowledge graphs (KGs) with large language models (LLMs) can enhance recommender systems (RSs) efficiency. By combining database and graph-based visualizations with natural language processing, our methodology deals with the complexities associated with recommending computer components. The results show an improvement in the system's capability to offer precise and contextual suggestions. KGs enable structured and interconnected data representation, which supports more refined reasoning and inference. Meanwhile, the LLM allows users to formulate queries in natural language with high interpretative accuracy. This combination is an advancement in KG-based RSs. We call this RS "Prometheus", a chatbot mainly designed for Lenovo internal use.

However, challenges remain. Building an extensive KG demands great efforts to refresh and ensure the accuracy of original data. In the future, we plan to broaden the reach of our KGs to contain more components and complex rules. This expansion will further refine the capability of the Prometheus chatbot. Furthermore, the potential to apply our methodology to other sectors offers promising research opportunities. Industries with complex inventories, such as automotive, electronics, or healthcare, could also potentially benefit from our approach. The broader application of integrating KGs and LLMs underscores the future possibilities for AI-driven RSs.

**6 Conclusion**

In conclusion, our work shows the benefits of KG collaborative LLM in improving RSs. Prometheus has proved useful in the industry for better computer components recommendations.

**7 Acknowledgments**

This project was supported by the Lenovo Desktop Computing Development Lab.

The codes are available on GitHub:

https://github.com/iamryanshengwang/Prometheus-Chatbot

The Lenovo database is not open access.

**Competing Interests**

The authors declare that they have no competing interests.

**Authors Contribution Statement**

Yunsheng Wang and Songhao Chen contributed equally to the conception and design of the study, data collection, analysis, and interpretation. Yunsheng Wang drafted the manuscript. Kevin Jin provided the grant and resources for the study. All authors reviewed and approved the final version.

**Ethical and Informed Consent for Data Used**

The data used in this study were sourced from Lenovo's proprietary databases. All necessary permissions were obtained from Lenovo to access and use this data for research purposes. As the data does not involve personal data, ethical approval and informed consent were not required.

**Data Availability and Access**

The codes used in this study are available on GitHub:

https://github.com/iamryanshengwang/Prometheus-Chatbot.

The Lenovo database used for constructing the knowledge graphs is not open access due to proprietary restrictions. Researchers interested in accessing the data can contact Desktop Computing Team Lead Songhao Chen: chensh8@lenovo.com for potential data sharing agreements.